# Corn Ear Detection and Orientation Estimation Using Deep Learning

Nathan Sprague[a], John Evans[b], Michael Mardikes[c]

[a]Purdue University, West Lafayette, 47907, IN, USA nspragu@purdue.edu

[b]Purdue University, West Lafayette, 47907, IN, USA jevansiv@purdue.edu

[c]Purdue University, West Lafayette, 47907, IN, USA mmardike@purdue.edu

**Abstract**

Monitoring growth behavior of maize plants such as the development of ears can give key insights into the plant's health and development. Traditionally, the measurement of the angle of ears is performed manually, which can be time-consuming and prone to human error. To address these challenges, this paper presents a computer vision-based system for detecting and tracking ears of corn in an image sequence. The proposed system could accurately detect, track, and predict the ear's orientation, which can be useful in monitoring their growth behavior. This can significantly save time compared to manual measurement and enables additional areas of ear orientation research and potential increase in efficiencies for maize production. Using an object detector with keypoint detection, the algorithm proposed could detect 90% of all ears. The cardinal estimation had a mean absolute error (MAE) of 18 degrees, compared to a mean 15 degree difference between two people measuring by hand. These results demonstrate the feasibility of using computer vision techniques for monitoring maize growth and can lead to further research in this area.

## 1. Introduction

Corn (*Zea Mays*) is an integral part of the global economy. Corn is a major food crop, especially in the United States, where over 88 million acres of corn were planted in 2018, and is continuing to grow (Saavoss et al. 2021). The yield of corn per acre has also been rapidly increasing, due to improved fertilizers, herbicides, genetics (Smith & Kurtz 2018), and planting (Licht et al. 2019). For this trend in yield to continue, further research must be conducted to better understand corn, including the development of the ears.

Traditional methods of measuring the orientation of corn ears are time-consuming and prone to human error. To address these challenges, this paper presents a computer vision-based system for detecting and tracking ears of corn in an image sequence. The proposed system uses deep learning techniques, such as object detection and regression, to accurately locate the ears and determine their orientation. The ability to accurately detect and track the ears can provide valuable information about their growth behavior and development, which can be used to improve yield and efficiency.

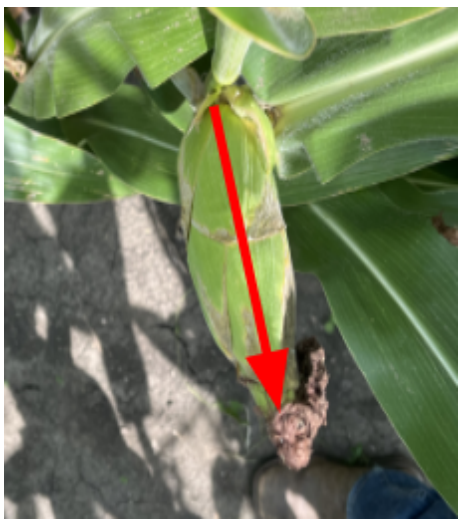

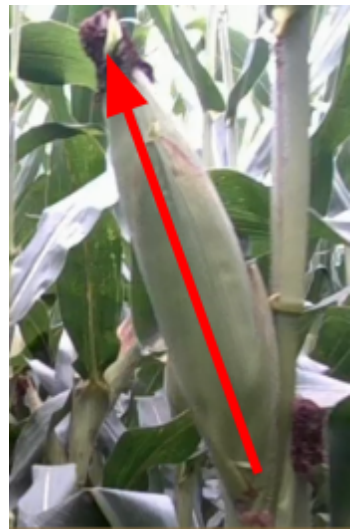

**Fig 1** Cardinal Ear Angle　　　　　　　　　　　　　　　**Fig 2** Ear angle off of stalk

The three-dimensional (3D) corn ear orientation can be broken down into a cardinal angle (Figure 1) and the angle off the stalk (Figure 2). The cardinal angle is the angle viewed from above, and points in a cardinal direction. It does not change much over the season. The cardinal angle is the most challenging angle to estimate because it varies between 0 – 360°, and is not naturally viewed by the camera. The angle off the stalk is the angle that is viewed from the side. It changes with moisture content.

The orientation of leaves in corn is a valuable indicator of its growth pattern and how it affects its neighboring plants. The presence and location of the leaves can have a substantial impact on the ear formation and overall yield (Subedi & Ma 2005). Tighter plant spacing can improve yield, but is limited by a variety of practical factors including nutrients (Li et al. 2020), and also sunlight absorption. When many corn plants are bunched together, they can impact the sunlight interception of the leaves of nearby plants. If the corn plants are properly oriented relative to each other, they can maximize exposure to sunlight (Toler et al. 1999), leading to better health and closer spacing, resulting in higher yield. The direction that a seed is placed can have an impact on the leaf orientation of the plant (Fortin & Pierce 1996), and can improve emergence rates (Torres et al. 2011). Seeds placed down and oriented in a certain pattern resulted in a predictable leaf growth orientation. The orientation of the ear may also play a role in the mechanical harvestability of the ear as the ear enters the combine (Mahmoud & Buchele 1975). By controlling the orientation of the ear, the yield may increase.

Since ears originate from a leaf, the orientation of the ear could be correlated to the leaf orientation, and therefore may be impacted by the seed placement as well. This theory has not been thoroughly studied, but an important step is being able to measure the ear orientation in the first place. The automated, high-throughput ear angle estimation in this project may help determine the ear orientation of many plants on a larger scale than what could be achieved with hand measurements.

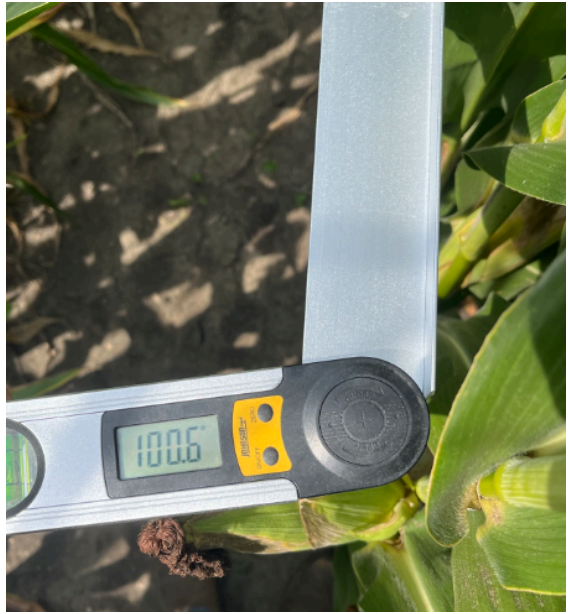

**Fig 3** Example of manual cardinal angle estimation using a digital angle finder.

The angle of the ear can be determined using a digital angle finder. To measure the cardinal angle, the person measuring tries to hold one end of the angle finder parallel to the row, and the other end in the direction of the ear (Figure 3). To measure the angle off the stalk, the angle finder is used in a similar way, but with one end of the angle finder pointed parallel to the stalk. This process is slow and prone to error. The plant is susceptible to torsion while being touched when measured. Pointing the end of the angle finder parallel to the row or stalk is challenging and the accuracy can vary between the people measuring. Therefore it is critical to develop a system capable of measuring rapidly and consistently.

Since the orientation that a seed is planted controls the orientation of the leaves of a corn plant, and ears originate from a leaf, then ears' angle may be influenced by seed orientation. To justify investing more resources into controlling the seed orientation, further study is needed. Due to the large number of ears in a field, measuring the orientation of all ears by hand is unfeasible, so a faster, automated method is necessary.

Determining the orientation of an ear on the stalk is a novel task. However, similar techniques have been used to detect ears and find other orientation characteristics of corn plants. One such study developed a model to estimate the height of ears on a corn plant (Wong 2019). It used a camera moving through a field pointed at ears at a fixed height. It would detect the ears using YOLOv3, and the detections were superimposed to a point pattern. All the detections were used in a ridge regression to estimate the average ear height per plot.

Object detectors have also been used to detect components of an ear as they enter combines. One such paper (Liu & Wang 2019) attached a camera strategically above the conveyor belt on a combine to detect broken ears. They used modified YOLOv3 and simulated images for the object detector to locate and classify the ears. Iowa State also performed a study with their robot to estimate the angle of leaves on a corn plant (Xiang et al. 2023). They used a custom stereo camera for RGB and depth. They estimated the 2D angle using three keypoints, one at the stalk, one at the leaf, and one at the node between the two. It would then use the stereo camera's depth to estimate the angle in three dimensions. This detection pipeline was limited to non-occluded leaves.

### 1.3. Objectives

The primary objective of this research is to develop a robust computer vision-based system for accurately localizing and estimating the 3D angles of maize ears within a production field. This involves creating a model that can identify and precisely measure the orientation of individual maize ears, thereby enabling farmers to make informed decisions regarding crop health and yield estimation. This can be achieved by the following objectives:

1. Evaluate the appropriate sensor system to record the ears in a field.
2. Develop an algorithm to locate the ears in the data.
3. Estimate the 3D pose of the detected ears.

4. Evaluate the model's accuracy for pose and detection ability.

## 2. Methods and Materials

### 2.1. Data Collection

An RGB depth camera was chosen to detect ears and their positions since ears are best distinguished from leaves by color and texture, which cannot be effectively captured by Lidar. Video data was acquired using the Open Source Connected Autonomous Rover (OSCAR) (Figure 4), a ground-based robot designed for non-destructive autonomous navigation through corn fields. OSCAR allows both sampling and imaging for any sensor attached. While its RTK GNSS capability allows point-to-point navigation outside the field, for in-row movement, it primarily relies on computer vision. OSCAR has differential steering on a 50.8 cm wheelbase, making it capable of easily maneuvering between standard 76.2 cm corn rows.

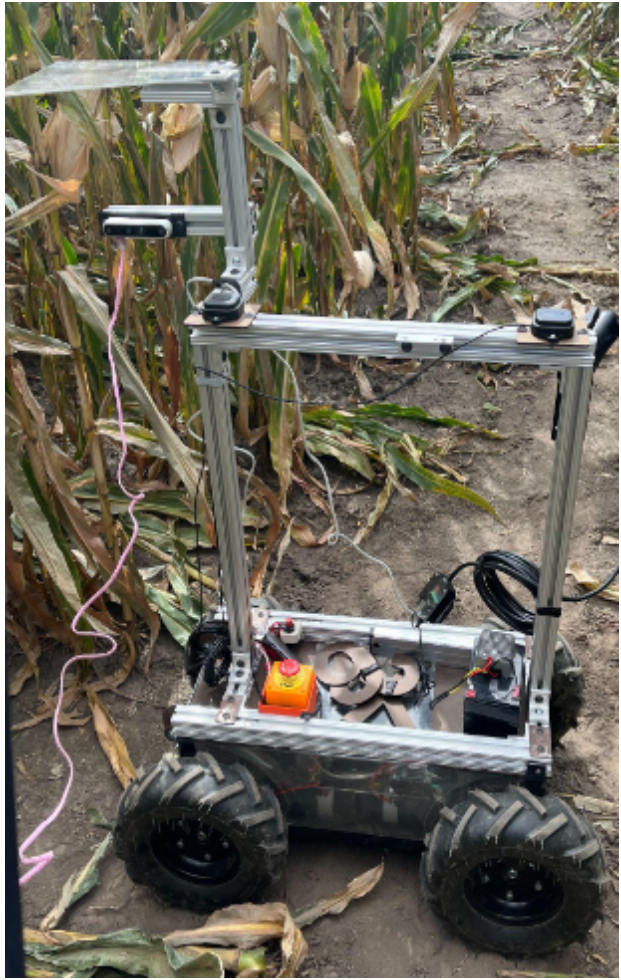

**Fig 4** OSCAR with camera mounted (top left). Note the acrylic sheet above camera to protect camera from leaves

To capture video for estimating ear orientation, a camera was placed on top of OSCAR, at the same level as most ears in the field, which was typically between 120-150 cm. Since leaves would often cover the camera from July to October, a clear acrylic sheet was placed above the camera to shield the camera from leaves. The Intel (Santa Clara, CA) RealSense and the Stereolabs (San Francisco, CA) Zed 2i were used to capture videos. Ears could be identified even in low resolution images, so videos were recorded at 720p.

### 2.2. Datasets

Approximately 50 hours of training video footage at ten frames per second (1.8 million total frames) of corn were taken from the robot traveling through the row. The footage for the training dataset was taken from various fields at Agronomy Center for Research and Education (ACRE), near West Lafayette, IN, from September 2022 to harvest time of 2023. Some depth footage was collected in addition to video footage using stereo cameras, but it quickly became apparent that the leaves from the canopy created too much noise for the depth to provide any useful footage. Five hours of testing footage was to determine the models' pose and detection accuracy was collected the same way, from the same fields as training data but in different rows. Labels were made by creating bounding boxes for all ears visible in the frame (Figure 5). Ears in the row closest to the camera were given one class, and ears from rows beyond were assigned a second class. Additionally, the pose of the ear was labeled by having two keypoints: one at the tip of the ear, and one at the node. The visibility of the keypoints were also labeled. Approximately 25,000 ears from 5,000 images were annotated this way for training and validation, and an additional 2,000 ears were annotated for testing.

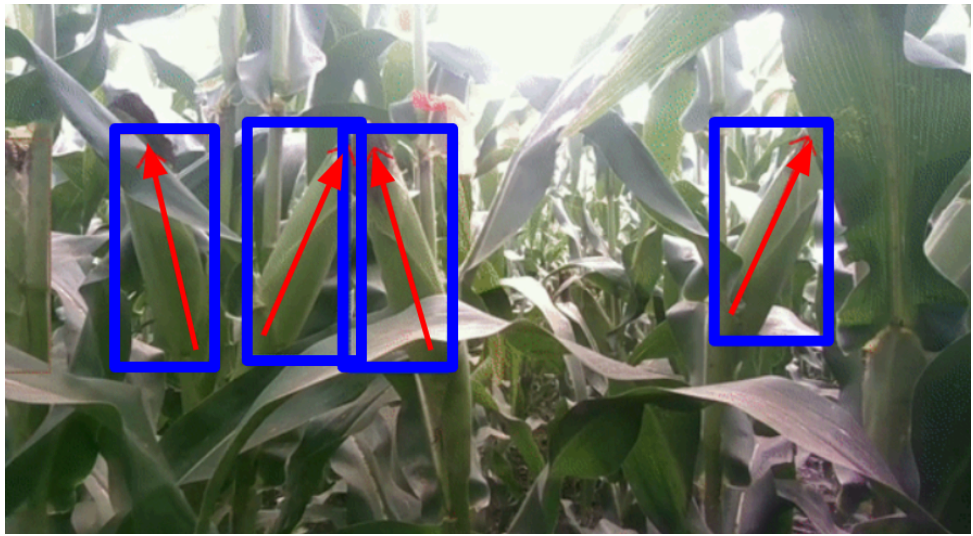

**Fig 5** Example ear annotation with bounding box and pose

The model's predictions were compared to 3D ground truth measurements collected using a digital angle finder. The measurements and videos were recorded in August 2023 in a field of Pioneer 1197AM corn. No training data was collected from this field. The cardinal angle was measured as degrees from north, ranging from 0–360°. The ear's angle off the stalk was measured so that if the ear were to point straight up, the angle would be 0°, and if the angle was pointing straight down, the angle would be 180°. Due to the torsional flexibility of the stalk and the organic shape of the ears, even hand measuring could not yield perfect results. To obtain a better understanding of the accuracy of the ground truth, two people measured each ear. As the corn plants begin to die and dry out, the ear's node connecting the ear to the plant weakens and the ear begins to point downward. Once the ears point straight down, the estimated accuracy of the cardinal ears with hand measurements was estimated to be ±45°. Every ten corn plants were marked off with a flag, which could be observed in the footage and counted off, to correlate the hand measurements to the video.

### 2.3. 3D Pose Estimation Pipeline

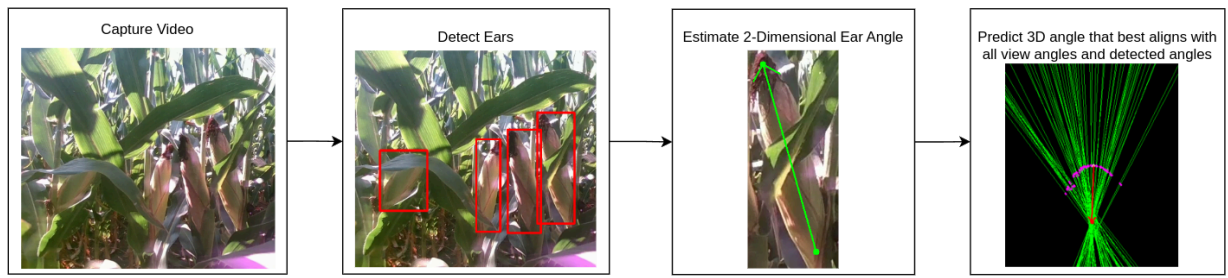

**Fig 6** Overview of angle estimation algorithm

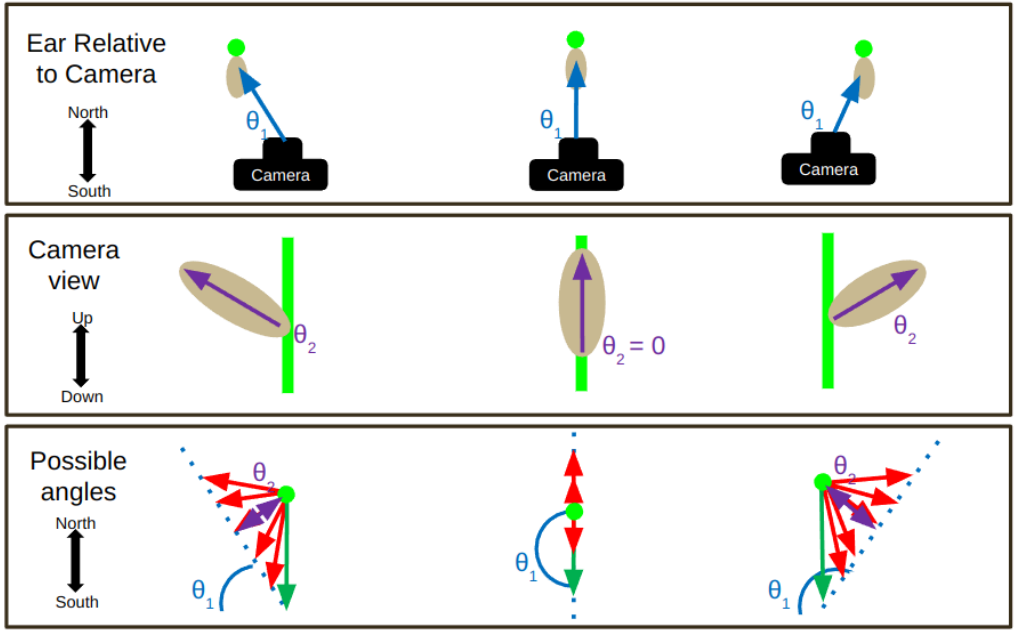

**Fig 7** Views used to calculate 3D angle

7a (top) angle of ear relative to camera

7b (middle) 2D angle of ear from camera perspective

7c (bottom) combined view

Estimating the ear's angle is a multi-step process (Figure 6). This method is able to estimate the 3D angle of the ears using RGB videos without depth by tracking ears and combining angle estimations. First, the ears are detected using an object detector. The x-coordinate of the detected ear is noted, as it corresponds to the angle of the ear relative to the camera (Figure 7a). Next, the 2D pose of the ear is estimated using a neural network (Figure 7b). This process is repeated for all frames of the video. Each ear is tracked across frames. Ear identified, all of the 2D estimated angles combined used to estimate the 3D angle. This algorithm was written in Python 3.11, with the OpenCV library for all image manipulations.

For each observation from the video, there are two 2D angles that can be observed. The first angle is the angle of the ear relative to the camera ($\theta_1$). This angle can be found using the x-position of the bounding box given by the object detector. This x-position can be converted to an angle by dividing it by the field of view of the camera. The second angle is the 2D angle pose of the ear found using keypoints as described earlier ($\theta_2$).

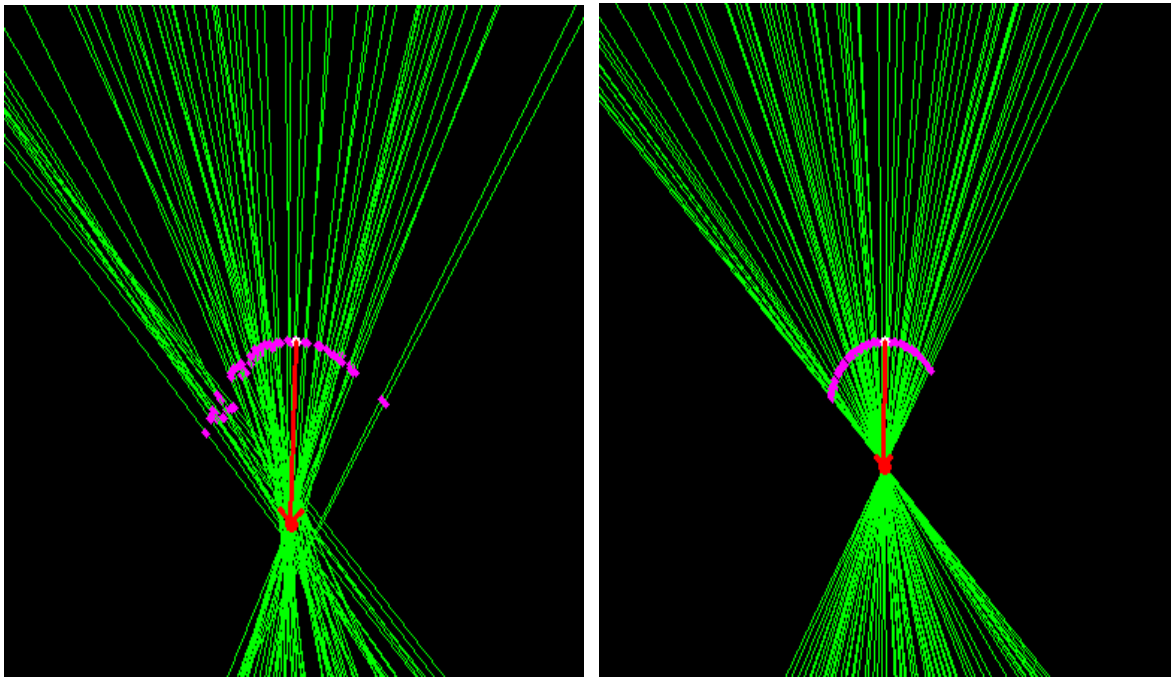

**Fig 8a** Final ear estimation by plotting:   **Fig 8b** Probability lines using ground truth

Probability lines for real ear angle (Green)

ear angle if no front/back angle (pink)

prediction angle (red)

Potential 3D angles can be plotted along a line (Figure 8). The angle of the green lines is the 2D pose of the ear, and the distance away from the center can be plotted as the distance from the center of the plot. If two or more observations are made at different camera angles, the lines of potential angles should intersect at one point. The angle of that intersection point relative to the center of the plot represents the cardinal angle of the ear, and the distance of the point from the center of the plot represents the ear's angle off of the stalk. This results in reasonable accuracy compared to the ground truth.

## 2.3. Object Detection

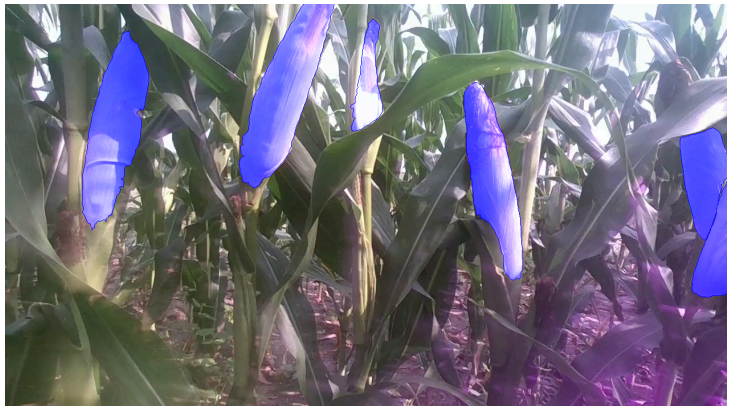

**Fig 9** Ears segmented using SAM

Two types of object detection model types were trained to detect ears, both using the YOLOv8. YOLOv8 (Varghese & M. 2024) is an object detector released in January 2023, while still performing in real-time. A segmentation model was trained using the YOLOv8's instance segmentation. To avoid labeling all images for segmentation by hand, Segment Anything Model (Kirillov et al. 2023) was used to generate training images, with the already labeled bounding boxes as prompts (Figure 9). A pose model was also trained using the labeled keypoints. A 85%/15% training/validation split was used. The models were trained for 100 epochs on the full dataset.

**2.4. 2D Pose Estimation**

The 2D pose of the ear was estimated using the segmentation and the keypoints.

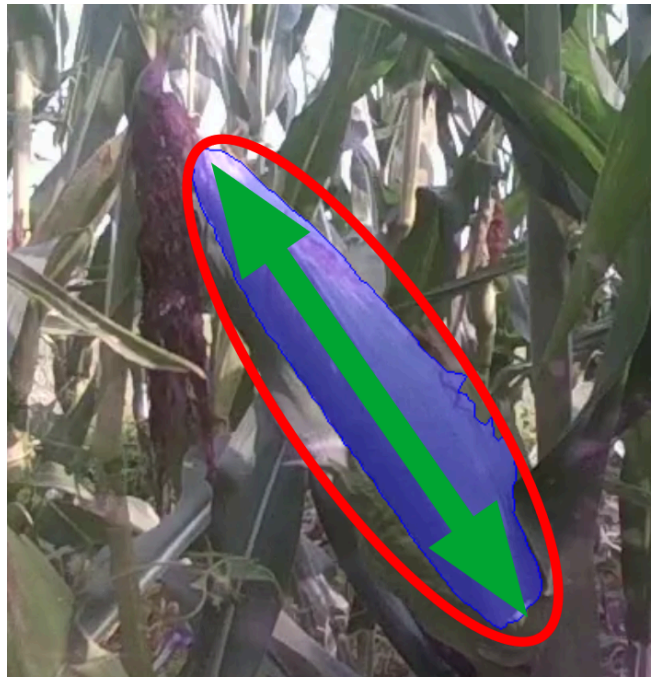

**Fig 10** Ear pose estimation from segmentation

The segmented ears typically are oval-shaped. An ellipse around the segmented ear using OpenCV's fit_ellipse() function (Figure 10). The angle of the major axis of the ellipse was used as the angle of the ear. Estimating the ear's angle using keypoints was performed by simply finding the relative angle between the two keypoints.

## 2.5. Tracking

Accurately estimating the 3-dimensional angle estimations heavily depends on multiple observations of the same ear, which requires proper tracking. Since each ear is observed multiple times, it is undesirable to count the same ear multiple times, resulting in an inaccurate estimate of the number of ears. Tracking ears of corn is particularly difficult, especially during mid-summer, when the leaves are upright and cause substantial occlusions to the camera. The final tracking technique employed was BoT-SORT (Aharon et al. 2022), built into the Ultralytics YOLO library. BoT-SORT is a real-time tracker that takes into account motion and appearance of detections. The first tracking technique described earlier could be used in combination with BoT-SORT to achieve even better tracking.

## 3. Results

## 3.1. Object Detector

The YOLOv8 object detector was able to reliably detect most ears in view. The ears that the object detector was unable to find were typically heavily obscured by leaves or other ears. The model's ability can be visualized by a precision-recall curve. This curve represents how accurate the model can be given the number of detections it can make. Ideally, this curve would be a straight line with a right angle at a precision of one and recall of one. The mean average precision (mAP) is the area under the curve. For the object detector, the mAP is 0.894 (Figure 12). For reference, the YOLOv9 AP on the COCO dataset is 0.556. The object detector trained for detecting ears performed better because it only needs to detect one class, while the COCO dataset has eighty. The confidence for detecting ears in the final pipeline was kept at the default 0.25, which allowed it to detect about 86% of all ears with around 90% accuracy. The specific accuracy of the object detector depended on the season and occlusions present.

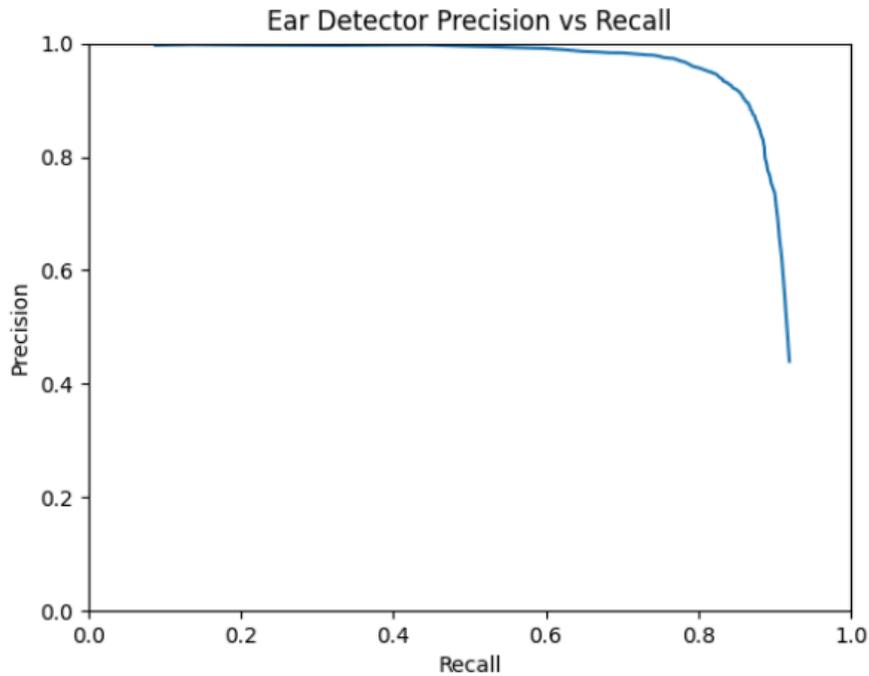

**Fig 12** Precision recall curve for YOLOv8 object detector

**3.2. 2D Angle Estimation**

The 2D pose of the ears were compared to the hand-labeled pose of the ears from the testing dataset. Because both the segmentation model and the keypoint model also localized the ears, the evaluation was performed by comparing angle of the detection with the highest confidence and an intersection over union (IOU) greater than 0.5.

The segmentation model had a MAE of four degrees and the keypoint method had a MAE of three degrees. Because the keypoint method provided better accuracy, the 3D angle estimates used in the remainder of this results section used the keypoints rather than segmentation.

## 3.3. Tracking

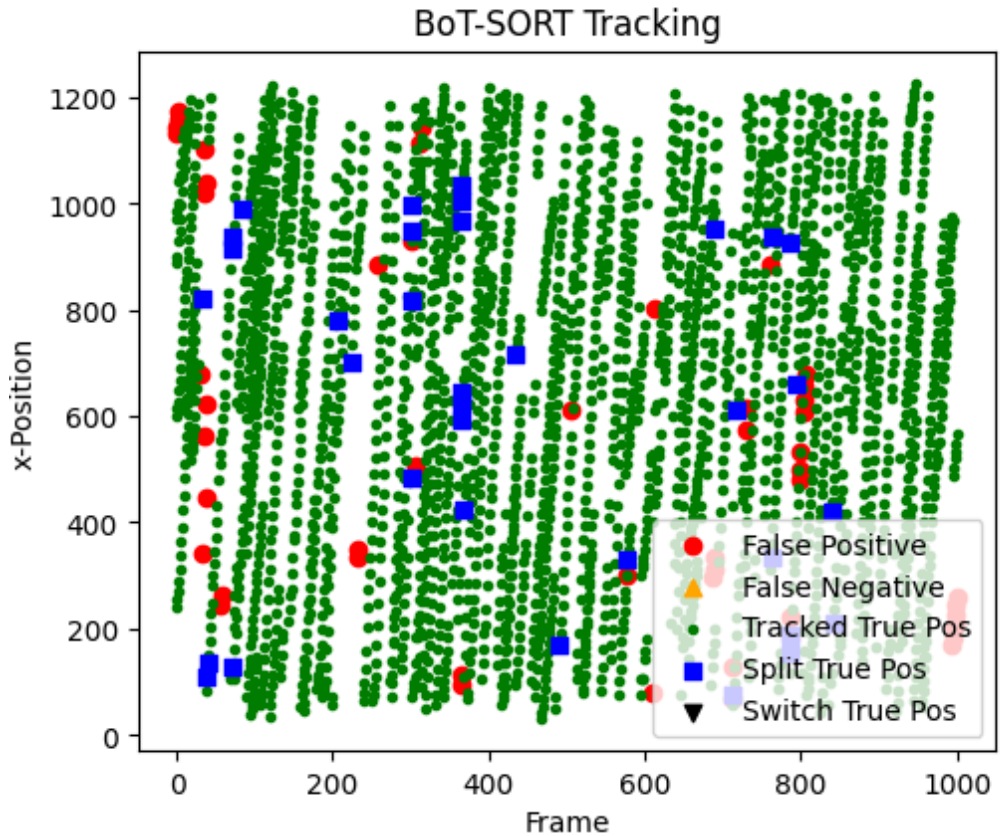

**Fig 13** BoT-SORT Tracking Accuracy

In a 100-ear hand-measured dataset from late August, the model made 36 tracking errors over the course of 2669 ear observations from 970 frames. While the error rate is low on a per-frame basis, each ear is observed in an average of 260 frames as the camera passes by. To properly observe an ear and estimate an angle, the ear must be tracked over every frame. The 36 tracking errors translated to 26 ears that were affected with tracking issues at least once out of the 100 ears (Figure 13). It is easy to fix the tracking errors by hand in post-processing, but the goal of this pipeline was to be fully automated. Nearly all tracking issues occurred when the ear was fully occluded for several frames and it came back into view. In areas where the ear is in view the entire time, the tracking performance is much better.

## 3.4. 3D Accuracy

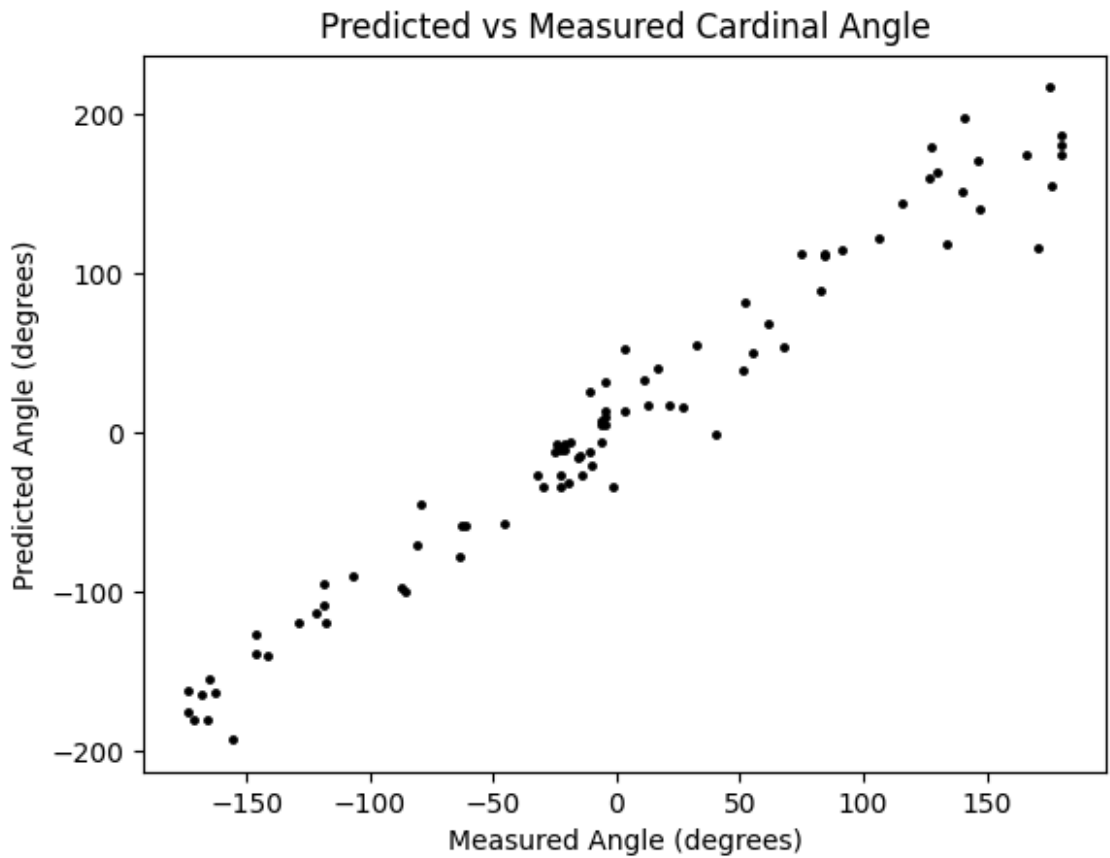

**Fig 14** Computer Vision Pipeline Accuracy

The same 100 hand-measured ear test segment used to evaluate the tracking was used to evaluate the model's 3D angle accuracy (Figure 14). To compare one-to-one with the ground truth, perfect tracking was assumed, where incorrect IDs were fixed to correspond to the correct detections. The ear angle estimations have a MAE of approximately 18° (r2=0.943), compared to a MAE of 12° between two people measuring the ears by hand.

### 3.5. Example Study: Temporal Ear Drop

One use for this pipeline is to determine the drop over time of the ears before harvest. As the moisture node connecting the ear to the stalk decreases, the ear's angle off the stalk typically drops. This can be easily observed qualitatively by simply looking at the ears, but can be challenging to quantify how this changes over time.

The same test segment of corn of 50 meters was regularly filmed between August 29th to harvest October 22nd, 2023. The model was applied and the angles and standard deviations were plotted (Figure 15). As empirically predicted, the ear's angle off the stalk dropped over time as the moisture content decreased. Most of the ears were upright (less than 90°) until mid-October, when the plant began to dry out and many ears dropped. By harvest, more than half of the ears were pointed straight down. The computer vision pipeline described a similar situation, showing the average ear angle off the stalk, and the standard deviation of the ear angles increasing as the season progressed.

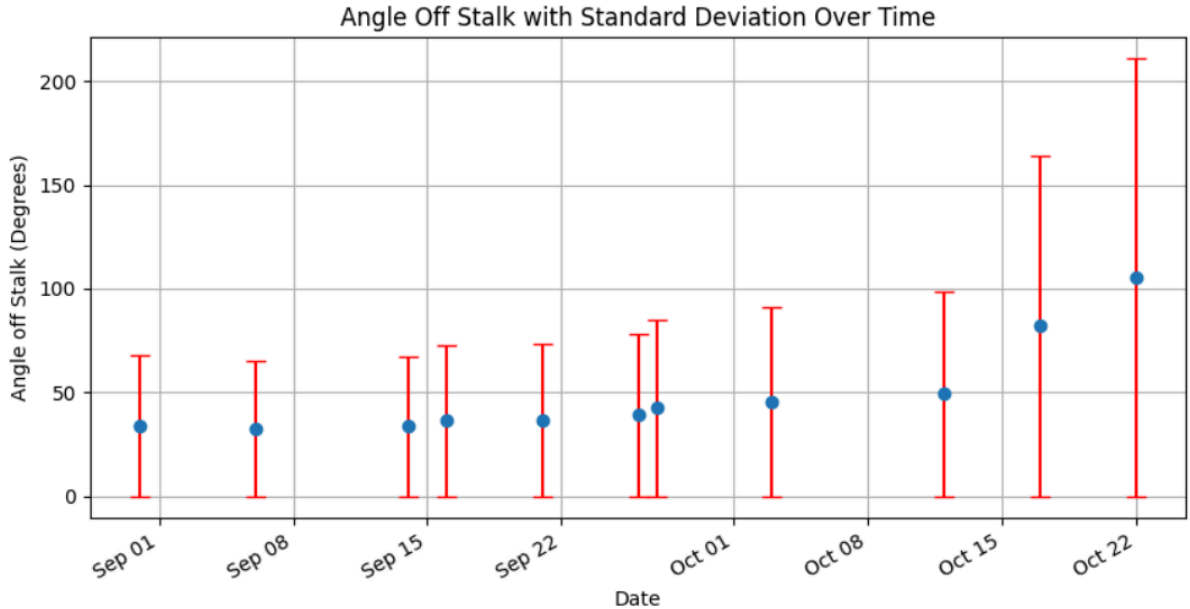

**Fig 15** Average angle off stalk and standard deviation over time

## 4. Conclusion

The orientation of the ear can provide useful insights into patterns in the growth and health of the plant. Determining the angle of the ear by hand is slow and prone to human bias. An automated detection model was developed to quickly and accurately estimate the ear orientation.

The pipeline developed detects stalks using an object detector, estimates the 2D pose of the ear using keypoints, and then observes the change in the angle of the ear as the camera moves by to estimate the 3D angle of the ear.

The accuracy of the object detector is very good in terms of differentiating between ears and leaves (MAP-50 = 0.894), but tracking remains challenging due to the frequent occlusions from leaves, even with the state-of-the-art tracking algorithm BoT-SORT. When the tracking errors are corrected by hand, the detection and pose estimation is accurate and can provide useful insights into the 3D orientation of ears in a field.